\documentclass[conference]{IEEEtran}
\IEEEoverridecommandlockouts
\ifCLASSINFOpdf
\else
\fi

\usepackage{amsmath,amssymb}
\usepackage{graphicx}

\usepackage{algorithm}
\usepackage{algpseudocode}
\algrenewcommand\algorithmicrequire{\textbf{Input:}}
\algrenewcommand\algorithmicensure{\textbf{Output:}}
\begin{document}

\title{CDKF-Track: Cluster-aware Data-Driven Kalman Filtering for Cooperative 3D Multi-Object Tracking
\thanks{This work has received funding from the EU’s Horizon Europe research and innovation programme in the frame of the AutoTRUST project “Autonomous self-adaptive services for TRansformational personalized inclUsivenesS and resilience in mobility” under the Grant Agreement No 101148123.}
}

\author{
\IEEEauthorblockN{Maria Damanaki$^{1,3}$, Nikos Piperigkos$^{1,4}$, Alexandros Gkillas$^{1,2}$, Aris S. Lalos$^{1,2}$}
\IEEEauthorblockA{$^1$Industrial Systems Institute, Athena Research Center, Patras Science Park, Greece\\
$^2$Omniloc.AI, Patras, Greece, $^3$Dpt. of Informatics \& Telecom., University of Ioannina, Arta, Greece, $^4$Avisense.AI, Patras, Greece\\
Emails: \{mdamanaki,piperigkos\}@isi.gr, \{gkillas,lalos\}@omniloc.ai 
}
}

\maketitle

\begin{abstract}
Multi-Object Tracking (MOT) is essential for EdgeAI perception systems, where accurate object localization and reliable identification enable safe decision-making. Single-agent MOT suffers from occlusions, sensor noise, and partial scene understanding in complex real-world scenarios. While multi-agent systems improve robustness by exploiting shared information, they introduce redundant measurements that lead to false data associations, and still struggle to capture nonlinear object dynamics. To address these challenges, we propose CDKF-Track, a Cluster-aware Data-Driven Kalman Filtering framework for Cooperative 3D MOT. The proposed method first fuses multi-vehicle 3D LiDAR detections through a Graph Laplacian-based formulation. Then, a cluster-aware redundancy reduction scheme groups spatially related detections and selects representative observations to reduce duplicate inputs to the tracker. The resulting detections are processed by a data-driven Kalman filter that learns object motion dynamics from data, reducing dependence on predefined linear motion assumptions. Furthermore, a wavelet-based temporal refinement module leverages the multi-resolution decomposition property of wavelets to attenuate short-term positional fluctuations and improve trajectory continuity. To the best of our knowledge, CDKF-Track is the first framework to jointly address detection-level fusion redundancy and learnable motion modeling in cooperative 3D MOT. Experimental results on the real-world V2V4Real dataset indicate that CDKF-Track achieves up to 27.99\% improvements in tracking accuracy over state-of-the-art multi-agent MOT methods.
\end{abstract}

\section{Introduction}

Recent advances in autonomous driving (AD) require high-performance perception algorithms to ensure reliable scene understanding and accident prevention. Single-Agent (SA) systems rely on onboard sensors, such as Cameras and LiDAR, and employ Deep Neural Networks (DNNs) to extract object-level representations, including centroids and bounding boxes associated with candidate physical objects. In dynamic driving scenes, these objects require to be accurately localized and consistently identified over time, forming the Multi-Object Tracking (MOT) scheme for situational awareness. However, SA perception remains constrained by limited sensor coverage, occlusions, and noisy observations. Connected and Autonomous Vehicles (CAVs) can mitigate these limitations by sharing complementary information across agents, and thus improving scene coverage. Nevertheless, Cooperative 3D MOT is affected by redundant observations, noisy trajectories, and complex motion patterns, which may lead to false associations, fragmented tracks, and identity switches. Hence, robust MOT is critical for reliable autonomous driving. 


Single-agent MOT follows either a tracking-by-detection or a joint detection-and-tracking paradigm to track object in the scene. 
In tracking-by-detection methods, a DNN-based detector first estimates 2D or 3D bounding boxes, which are subsequently associated with trajectories to initialize new tracks or update object states using Kalman or Bayesian filtering models \cite{ab3dmot,9562072,9810346, arioka2026adaptive,10164676}.
These methods are computationally efficient and scalable across datasets, as detection and tracking can be trained or optimized independently. In contrast, joint detection-and-tracking approaches integrate object detection and temporal association within a unified DNN-based framework \cite{9636311, 8575355,lu2020retinatrack}, requiring higher computational cost and large-scale training data. 
Despite the advances, many MOT approaches rely on predefined motion models, such as classical or extended Kalman filtering limiting tracking performance under complex real-world object kinematics. Data-driven Kalman-filtering approaches \cite{wang20243d, hybridtrack} constitute a promising direction for robust MOT by learning object motion dynamics directly from data and reducing reliance on predefined motion and noise assumptions.
In addition, trajectory refinement based on previous timestamps can improve temporal consistency, such as wavelet-based methods to attenuate high-frequency positional noise \cite{fard2017new}.
Nevertheless, SA MOT is inherently constrained motivating the transition toward Cooperative MOT ensuring high situational awareness.

Existing cooperative tracking methods can be broadly categorized into \textbf{early, cascade, and joint/end-to-end} multi-agent MOT, depending on the stage at which multi-vehicle information is exploited. In \textbf{early cooperative tracking}, detections from multiple agents are fused before trajectory association, resulting in a simple and computationally efficient scheme. For instance, V2V4Real \cite{v2v4real} aggregates bounding boxes from two CAVs and employs AB3DMOT \cite{ab3dmot} for tracking. Similarly, graph-aware methods refine multi-agent detections through structured fusion mechanisms \cite{tsa}, while uncertainty-aware approaches propagate detection uncertainty into the Kalman filtering and association stages \cite{10430224}. In \textbf{cascade cooperative tracking}, detections from different agents are introduced sequentially into the tracking pipeline \cite{10588576, chiu2024probabilistic}. However, prioritizing a specific agent may bias the association process under noisy or unreliable observations. In \textbf{joint/end-to-end cooperative tracking}, detection, association, and tracking are integrated within a unified framework \cite{10148929, zhong2025cooptrack}.
Although unified end-to-end approaches are promising, they often rely on feature-level representations, dedicated training procedures, or sensing configurations that differ from LiDAR-based vehicle-to-vehicle (V2V) tracking-by-detection pipelines. In this context, detection-level cooperative MOT remains an important direction, as it preserves modularity, interpretability, and compatibility with existing 3D detection outputs.

Overall, cooperative MOT is affected by redundant multi-agent detections and insufficient motion modeling under complex real-world dynamics. When multi-vehicle observations are fused, several bounding boxes may correspond to the same physical object, increasing false positives and complicating trajectory association. 
Although uncertainty-aware and end-to-end MA MOT schemes can reduce some of these errors, they often depend on additional uncertainty modeling, dedicated training, or feature-level information exchange. Therefore, detection-level MA MOT requires interpretable mechanisms to reduce duplicate detections while preserving compatibility with existing 3D detection pipelines. Accordingly, cluster-aware fusion, data-driven Kalman filtering, and temporal refinement jointly provide a promising scheme for MA MOT by reducing duplicate detections, modeling nonlinear object kinematics, and smoothing short-term positional fluctuations.


Therefore,  we propose \textbf{a novel Cluster-aware Data-Driven Kalman Filtering, (CDKF-Track)} scheme that jointly addresses multi-agent detection redundancy and nonlinear motion modeling in a unified tracking-by-detection pipeline. First, multi-agent 3D detections are fused through a fully connected graph topology based on the Graph Laplacian operator. Then, a cluster-aware redundancy reduction module groups spatially related detections, where each cluster corresponds to a potential physical object and a representative detection is selected to reduce duplicate observations and false associations. The resulting representative detections are associated with existing trajectories and processed by a data-driven Kalman filtering tracker, which learns object motion dynamics from data and reduces reliance on predefined linear motion assumptions. Finally, a wavelet-based temporal refinement module exploits historical trajectory information to smooth short-term positional fluctuations and improve trajectory continuity. To the best of our knowledge, CDKF-Track is the first framework to jointly address detection-level fusion redundancy and learnable motion modeling in cooperative 3D MOT.

Overall, the main contributions can be summarized as: 
\begin{itemize}
\item We design a data-driven multi-agent scheme to precisely estimate the position of surrounding objects in the scene capturing non-linear kinematics.
\item We formulate a cluster-aware Graph Laplacian fusion pipeline that refines multi-agent detections and reduces duplicate observations before trajectory association.
\item We develop a wavelet-based scheme exploiting the historical information to smooth the trajectories in time. 
\item An extensive experimental evaluation has been conducted on V2V4Real \cite{v2v4real}, the first large-scale real-world V2V dataset, where the CDKF-Track achieves significant improvements up to \textbf{27.99\%} in tracking accuracies against state-of-the-art Cooperative MOT schemes.
\end{itemize}

\section{System model and Preliminaries}
\label{preliminaries}
This section presents the system model and the preliminaries required to describe the proposed method.
\subsection{System Model}
\label{systemmodel}
In this framework, we adopt the tracking-by-detection paradigm for multi-agent detection fusion and tracking, where multiple CAVs share their 3D detections and project them to a central agent. At timestamp $t$, 3D detector is applied independently to the LiDAR point cloud of each agent $i$ obtaining $N_{D^i}$ 3D detections noted by $\mathcal{D}^{(i,t)} \in \mathbb{R}^{N_{D^i} \times 7}$, and each $m$ observation is described by $\boldsymbol{x_D^{(i,m,t)}}={[x_{i,m},y_{i,m},z_{i,m},h_{i,m},w_{i,m},l_{i,m},\theta_{i,m}]}^T$ with $x_{i,m},y_{i,m},z_{i,m}$ its centroid, $h_{i,m},w_{i,m},l_{i,m}$ its height, width, length, and $\theta_{i,m}$ its rotation angle. In this framework the central agent corresponds to ego-vehicle. 
Furthermore, the $N_I$ active tracks at timestamp $t$ form the set $\mathcal{I}^{(t)} \in \mathbb{R}^{N_I \times 7}$, where $\boldsymbol{x_I^{(k,t)}}={[x_k,y_k,z_k, h_k,w_k,l_k,\theta_k]}^T$ is the track state of the $k$ track described by its centroid $(x_k,y_k,z_k)$, dimensions $(h_k,w_k,l_k)$ and rotation angle $(\theta_k)$. It is noted that the tracks are considered in the common scene so they are not specified by any agent.


The tracking stage adopts a learnable Kalman-Filtering formulation inspired by HybridTrack \cite{hybridtrack}, a state-of-the-art SA MOT tracker. 
For each $k$ active track, the posterior state at timestamp $t-1$ is denoted as $\boldsymbol{\hat{\mathbf{x}}_{I}^{(k,t-1)}}\in\mathbb{R}^{7}$. Instead of relying on a predefined linear transition matrix, the data-driven tracker predicts a transition residual from the previous state and the ${\mathcal{H}_{k}^{(t-1)}} \in \mathbb{R}^{W_k \times 7}$ historical states of the track considering $W_k$ previous timestamps:
\begin{align}
\boldsymbol{\Delta \mathbf{x}_{I}^{(k,t)}}
=
f_{\theta}\left(
\boldsymbol{\hat{\mathbf{x}}_{I}^{(k,t-1)}},{\mathcal{H}_{k}^{(t-1)}}
\right),
\label{eq:learned_residual}
\end{align}
where $f_{\theta}(\cdot)$ is a learned transition residual predictor. Hence, the prior state is obtained as:
\begin{align}
\boldsymbol{\bar{\mathbf{x}}_{I}^{(k,t)}}
=
\boldsymbol{\hat{\mathbf{x}}_{I}^{(k,t-1)}}
+
\beta_{k}^{t}\boldsymbol{\Delta \mathbf{x}_{I}^{(k,t)}},
\label{eq:hybridtrack_prior}
\end{align}
where $\beta_{k}^{t}$ is a scaling factor used to stabilize the prediction during early track initialization or missed-detection intervals. 

When a new measurement $\boldsymbol{\mathbf{z}_{I}^{(k,t)}}$ of the $k$ track arrives, track's posterior state is updated through a Kalman-like correction:
\begin{align}
\boldsymbol{\hat{\mathbf{x}}_{I}^{(k,t)}}
=
\boldsymbol{\bar{\mathbf{x}}_{I}^{(k,t)}}
+
\mathcal{G}_{k}^{(t)}
\left(
\boldsymbol{\mathbf{z}_{I}^{(k,t)}}
-
\mathbf{H}\boldsymbol{\bar{\mathbf{x}}_{I}^{(k,t)}}
\right),
\label{eq:learned_kalman_update}
\end{align}
where $\mathcal{G}_{k}^{(t)}$ is a learned Kalman gain estimated from data and $\mathbf{H}=\mathbb{I}_7 \in \mathbb{R}^ {7\times 7}$ is the measurement matrix with the $\mathbb{I}_{7}$ identity matrix. This formulation preserves the recursive structure of Kalman filtering while replacing hand-crafted transition and noise assumptions with learnable components.


\subsection{Graph Signal Processing and Laplacian Preliminaries}
\label{graph_preliminaries}
The Graph Laplacian operator is a fundamental tool in Graph Signal Processing for reconstructing graph signals from differential coordinates and anchor points in a least-squares sense \cite{sorkine2005laplacian}. In the proposed framework, this formulation is adopted to fuse and refine multi-agent detections within a fully connected graph, where spatial relationships among observations are encoded through the graph topology following a similar graph-based formulation as proposed in \cite{tsa}. This tool considers at time $t$ an undirected graph $J^{t} = (\mathcal{V}^t, \mathcal{E}^t)$, where $\mathcal{V}^t$ and $\mathcal{E}^t$ denote the set of with $N^t$ vertices and edges, respectively. It derives the Laplacian matrix $\boldsymbol{L^{(t)}}$ to formulate the links of each node and is equal to $\boldsymbol{L^{(t)}=D^{(t)}-A^{(t)}}$, where  $\boldsymbol{D^{(t)}}$ , $\boldsymbol{A^{(t)}} \in \mathbb{R}^{{N^t}\times{N^t}}$ are the well-known degree and adjacency matrices. Furthermore, the differential vector $\boldsymbol{\delta^{t}} = [\delta^{(1,t)} \hdots \delta^{(N_t,t)}]$, captures the inherent spatial geometries, where $\delta^{(i,t)}= \sum_{i}^{{N^t}}({v^{(i,t)}}-{v^{(j,t)}})$ among all connected vertices. The anchor vector $\boldsymbol{c^{(t)}}$ serves as the complementary information of the nodes. In order to formulate a least-squares optimization problem, the Laplacian matrix is extended with the identity matrix in order to avoid singularity issues and thus is $\boldsymbol{\tilde{L}^{(t)}=[L^{(t)} \mathbb{I}_{N^t}]^T}$. The differential vector and anchor points detail the measurement vector $\boldsymbol{\tilde{\delta}^{(t)}}$. Hence, the objective is the minimization of the cost function: 
\begin{equation}
    argmin_{\boldsymbol{v^{(t)}}} ||{\boldsymbol{\tilde{L}^{(t)}v^{(t)}-\tilde{\delta}^{(t)}}}||^2_2
\end{equation}
with unique analytical solution 
\begin{equation}
    \boldsymbol{v_{*}^{(t)}} = (\boldsymbol{(\tilde{L}^{(t)})^T\tilde{L}^{(t)})^{-1}(\tilde{L}^{(t)})^{T}}\boldsymbol{\tilde{\delta}^{(t)}}, \ \ \ \boldsymbol{v_{*}^{(t)}} \in \mathbb{R}^{N^t}
\end{equation}
This tool will provide the basis to fuse multi-agent detections by formulating as graph nodes their spatial attributes.

\section{Cluster-Aware Data-Driven Tracking Framework}
\label{proposed_methods}

\begin{figure*}[ht]
\centering
 \includegraphics[scale=0.35]{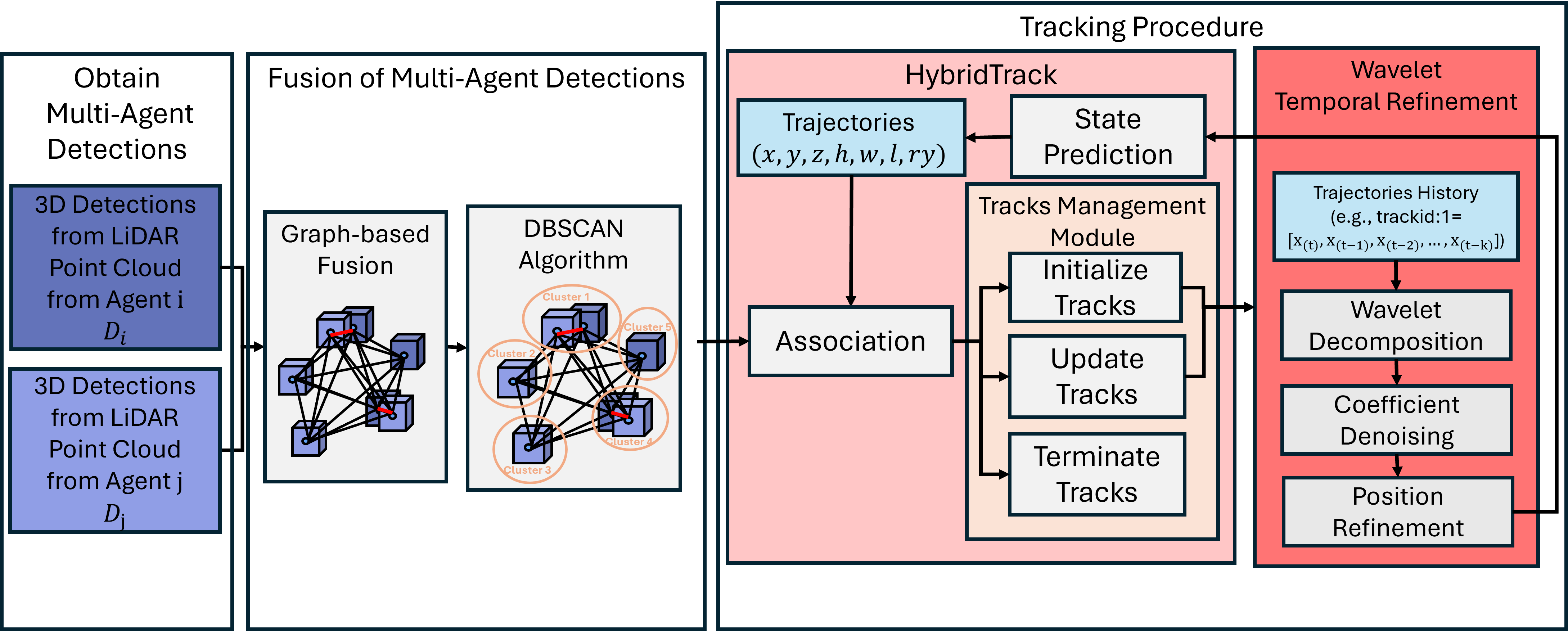}
 \caption{Data-Driven Multi-Agent Multi-Object Tracking Scheme. Detections from each agent formulate a fully connected graph topology in order to fuse multi-agent information and a clustering algorithm is applied to cut the redundant information, where each cluster corresponds to potential real object of the scene. The representative of each fused and refined cluster is associated with existing trajectories where their states are predict and updated through a data-driven model based tracker localizing precisely the bounding boxes. Finally, a wavelet-based transformation is developed to smooth each trajectory separately smoothing further the position in time considering historical information and thus deprecate false associations in the next timestamps.}
  \label{fig:scheme}
\end{figure*}

We propose \textbf{a novel Cluster-aware Data-Driven Kalman Filtering, (CDKF-Track)} framework for Cooperative 3D MOT. The proposed scheme follows an early tracking-by-detection MA MOT and aims to address two main limitations of cooperative tracking: redundant multi-agent detections and limited motion modeling under complex object dynamics. Specifically, the pipeline consists of two main modules. The first module performs graph-aware multi-agent detection fusion and cluster-aware redundancy data reduction. The second module performs data-driven cooperative tracking using a learnable Kalman filtering, followed by wavelet-based temporal refinement.
A general overview of the CDKF-Track, as depicted in Fig. \ref{fig:scheme}, is: i) firstly, 3D multi-vehicle detections are fused through the Graph Laplacian Operator and anchor points formulating a fully connected graph topology, ii) the DBSCAN clustering algorithm is adapted to identify overlapped information and reduce data redundancy by defining representatives for each cluster which correspond to potential real object, iii) the cluster representatives are associated with existing trajectories and used as measurements to update track states through the data-driven Kalman filter. The learnable filter aims to localize precisely the objects in the scene capturing non-linear motion without assuming constant velocity or in generally linear motion models. iv) Finally, the proposed method considers also a wavelet-based transformation to attenuate short-term positional fluctuations of each trajectory improving tracks continuity. Overall, the proposed method tracks objects in the scene by addressing nonlinear motion dynamics and redundant multi-agent observations, providing a practical pipeline for real-world Cooperative MOT scenarios.

\subsection{Cluster Graph-Aware Multi-Agent Detections Fusion}
This module fuses multi-vehicle detections through a Graph Laplacian-based formulation and adopts a cluster-aware paradigm to suppress overlapping observations before tracking, thus mitigating false data associations. The graph-aware scheme combines multi-agent observations using differential coordinates and anchor constraints. Then, DBSCAN groups spatially related detections based on centroid distances, where each cluster is treated as a potential physical object and one representative detection is selected. Additionally, the confidence scores of overlapping detections are boosted, since agreement among multiple observations indicates a higher likelihood of corresponding to a real object.


First, a fully connected graph is constructed over the multi-agent detections using the Graph Laplacian formulation described in Section~\ref{graph_preliminaries}. The nodes of the graph $\mathcal{J}^{(t)}$ correspond to the spatial attributes of the multi-vehicle detections, while the edges connect observations both within the same agent and across different agents, forming a fully connected topology. Although the current setting considers two CAVs, denoted by $i$ and $j$, the formulation can be extended to a larger number of agents. The multi-vehicle detections of agent $i$ and $j$ are associated using the 3D Complete Intersection over Union (3D CIoU) \cite{zheng2021enhancing} metric and the Hungarian Algorithm (HA). Matched detections are assumed to correspond to the same physical object and are used to define anchor constraints in the Graph Laplacian refinement process.

After the graph-aware fusion, more than one detections may correspond to the same observed object resulting to an increased number of false positive detections. Hence, the DBSCAN clustering algorithm is adopted as an id-agnostic clustering without requiring data uniform distribution to classify objects based on the 3D Euclidean Distance on $x,y,z$ spatial attributes.  Each cluster is considered as a potential real object which may consist more than one detections, and thus a representative is defined to cut the redundant information.
The bounding box with the highest confident score $cs_d$ of each cluster is defined as the representative and in case of overlapped detections, the confidence score is boosted by $cs_d' = 1- (1-cs_d)^2$ as it is more likely to represent a real object. These fused and refined detections are passed to the data-driven tracking pipeline in order to localize and identify the objects over time. 

\subsection{Data-Driven Multi-Agent Multi-Object Tracking}

This module tracks objects over time using a learnable Kalman-filtering formulation to model non-linear object dynamics, followed by a wavelet-based temporal refinement stage to smooth trajectory fluctuations without introducing an additional trainable component. The data-driven Kalman filter updates object states from associated representative detections, while the wavelet module exploits recent trajectory history to attenuate short-term positional variations. Together, these components improve temporal consistency and tracking robustness in real-world scenarios characterized by complex motion patterns, noisy detections, and incomplete observations.

The tracking procedure is performed at each timestamp following a tracking-by-detection paradigm. Specifically, the refined representative detections obtained from the cluster-aware graph fusion module are passed to the tracking stage and associated with existing trajectories using the 3D CIoU and the HA. Unmatched detections are used to initialize new tracks, while unmatched trajectories are propagated to the next timestamp as long as their inactive $age$ does not exceed the predefined threshold $th_a$. Matched trajectories update their states using the corresponding detections according to the data-driven Kalman-filtering formulation as described in Eq. \ref{eq:learned_kalman_update}.

After the association and state-update stage, a wavelet-based temporal refinement module is adopted to smooth short-term positional fluctuations of active trajectories. For each active trajectory $k$ and $x_k$ the spatial coordinate on $x$-axis, a temporal signal $\mathbf{x}_{k}^{(W_w)}$ is constructed from the recent trajectory history and decomposed through multi-level Haar wavelet transforms, yielding approximation and detail coefficients $\{a_{k}^{(\ell)}, q_{k}^{(\ell)}\}$ at each decomposition level $\ell$. The detail coefficients are denoised through noise-adaptive soft-thresholding with threshold
\begin{equation}
\lambda^{(\ell)}
=
\alpha \hat{\sigma}^{(\ell)}
\sqrt{2\log N_{\ell}},
\end{equation}
where $\alpha$ denotes the wavelet threshold scale, $N_{\ell}$ is the number of detail coefficients at level $\ell$, and $\hat{\sigma}^{(\ell)}$ is a median absolute deviation (MAD)-based noise estimate \cite{donoho1994ideal}. This process attenuates high-frequency positional oscillations while preserving the dominant motion structure of the trajectory. The refined temporal signal $\widetilde{\mathbf{x}}_{k}^{(W_w)}$ is subsequently reconstructed through inverse Haar synthesis and propagated to the next timestamp, improving trajectory smoothness and temporal continuity.

Overall, the proposed method combines graph-aware multi-agent fusion, cluster-based redundancy reduction, data-driven motion modeling, and temporal trajectory refinement to improve cooperative 3D MOT robustness under complex real-world driving conditions and is summarized in \textbf{Algorithm \ref{alg:proposed_framework}}.

\begin{algorithm}
\caption{\textbf{:Cluster-aware Data-Driven Kalman Filtering (CDKF-Track)}}
\label{alg:proposed_framework}
\begin{algorithmic}[1]
\Require Multi-agent detections $\mathcal{D}^{(i,t)}, \mathcal{D}^{(j,t)}$ and existing tracks $\mathcal{I}^{(t)}$
\Ensure Updated tracks $\mathcal{I}^{(t+1)}$
\For{each timestep $t=1,\ldots,T$}
    \State Project detections from all agents to the ego-vehicle coordinate frame.
    \State Construct a graph over multi-agent detection centroids.
    \State Refine detection positions using the Graph Laplacian (Section~\ref{graph_preliminaries}).
    \State Cluster refined detections and select one representative per cluster.
    \State Associate representative detections with existing tracks.
    \State Update matched tracks; initialize new tracks from unmatched detections.
    \State Predict unmatched tracks if inactive age is below $th_a$.
    \State Apply wavelet-based temporal refinement to smooth track positions.
\EndFor
\end{algorithmic}
\end{algorithm}

\section{Experimental results}
\label{experiments}
This section evaluates CDKF-Track against single-agent and multi-agent MOT baselines on the V2V4Real \cite{v2v4real} dataset. 
\subsection{Implementation Details}
\subsubsection{Dataset} We evaluate the proposed framework on V2V4Real \cite{v2v4real}, the first large-scale real-world V2V perception dataset with 2 simultaneously driven vehicles, Tesla (ego-vehicle) and Astuff. This dataset consists of 32 driving training sequence and 9 testing sequences. We utilize the 3D PointPillar \cite{8954311} detector and for the MA MOT we project all the observations to Tesla's Coordinate System.
\subsubsection{Training} 
The training of the data-driven HybridTrack \cite{hybridtrack} was performed on the V2V4Real dataset for the "Car" class using trajectory-level annotations derived from the dataset's tracking training split. The model was optimized using Adam with a learning rate of $10^{-3}$, weight decay of $10^{-5}$, and batch size 128. The training objective was formulated as a composite L1-based loss over both the predicted 3D bounding-box state and the Kalman prior state, accounting for position, dimensions, and yaw estimation errors. 
\subsubsection{Relevant Approaches} We compare our framework against the state-of-the-art MA MOT, the V2V4Real \cite{v2v4real}, DMSTrack \cite{chiu2024probabilistic}, and TSA Graph Lap-CoMOT \cite{tsa} and the SA AB3DMOT \cite{ab3dmot} tracker on each CAV. In the V2V4Real, detections from all agents are aggregated and forwarded to AB3DMOT \cite{ab3dmot} for tracking. DMSTrack is a deep learning-based method that leverages detector backbone features to quantify detection uncertainty, prioritizing ego-vehicle observations. TSA-Graph Lap-CoMOT constitutes a graph-based fusion framework that performs two-stage tracking association. 
\subsubsection{Evaluation Metrics} We employ the well-known evaluation metrics as they proposed in 
\cite{ab3dmot} including sAMOTA, AMOTA, AMOTP, mostly tracked (MT) (i.e., the percentage of correct tracking of objects over the 80\%
of their life), with respect to True Positive (TP), False Positive (FP), False Negatives (FN), and Identity Switches (IDSW).

\subsection{Evaluation Results}
\begin{table*}[ht]
\centering
\caption{Performance comparison on the V2V4Real dataset. Plus (minus) sign indicate the improvement relative to the best-performing state-of-the-art multi-agent MOT baseline. The proposed framework consistently demonstrates superior object tracking performance by effective handling non-linear motions and reducing redundant information.}
{ 
\begin{tabular}{|c|c|c|c|c|}
\hline
\textbf{Methods} & \textbf{sAMOTA (\%)$\uparrow$} & \textbf{AMOTA  (\%) $\uparrow$} & \textbf{AMOTP  (\%) $\uparrow$} & \textbf{MT (\%) $\uparrow$}\\
\hline
\textbf{Tesla (SA MOT) \cite{ab3dmot}} & 52.42 & 18.6 & 39.88 & 45.02\\
\textbf{Astuff (SA MOT)\cite{ab3dmot}} & 54.85 & 20.86 &  35.55  & 48.56\\
\textbf{V2V4Real (MA MOT)\cite{v2v4real}} & 82.09 & 40.51 & 59.79 & 68.05\\
\textbf{DMSTrack (MA MOT)\cite{chiu2024probabilistic}} & 84.02 & 42.11 & 57.01 & 65.07\\
\textbf{TSA-Graph MOT (MA MOT) \cite{tsa}} & 83.62&42.44&61.64&71.16\\ 
\textbf{CDKF-Track (MA MOT)(ours)} & \textbf{85.07(+3.63\%)}&\textbf{43.17 (+6.57\%)}&\textbf{65.39 (+14.62\%)} &\textbf{83.28 (+27.99\%)}\\
\hline
\end{tabular}
}
\label{tab:general}
\end{table*}

\begin{table}[ht]
\centering
\caption{Ablation Study: Performance comparison on the V2V4Real dataset}
\resizebox{9cm}{!}{ 
\begin{tabular}{|c|c|c|c|c|}
\hline
\textbf{Components of CDKFTracker} & \textbf{sAMOTA $\uparrow$} & \textbf{AMOTA$\uparrow$} & \textbf{AMOTP $\uparrow$} & \textbf{MT $\uparrow$}\\
\hline
\textbf{HybridTrack only} & 64.76 & 33.09 & 64.9 & \textbf{83.9} \\
\textbf{DBSCAN+HybridTrack} & 82.16  & 42.41 & \textbf{66.07}  & 82.07\\
\textbf{Graph+DBSCAN+HybridTrack} & 82.68 & 42.95 & 65.25 & 82.07\\
\textbf{Graph+DBSCAN+HybridTrack+Wavelet} & \textbf{85.07} & \textbf{43.17} & 65.39 & 83.28\\
\hline
\end{tabular}
}
\label{tab:ablation_components}
\end{table}

Table \ref{tab:general} demonstrates the average tracking performance across all V2V4Real testing sequences. The \textbf{CDKF-Track} outperforms among the state-of-the-art MA MOT methods, with improvements up to \textbf{3.63\%} and \textbf{6.57\%} in sAMOTA and AMOTA, by reducing the redundant information through the cluster graph aware pipeline addressing the increased number of FP and IDSW-related tracking accuracy errors. Additionally, the proposed method captures the non-linear objects dynamics enhancing tracking precision up to \textbf{14.62\%} in AMOTP. Furthermore, outstanding performance is achieved up to \textbf{27.99\%} in MT by consistently tracks objects in the scene for over of 80\% of their lifetime. Therefore, the \textbf{CDKF-Track} achieves superior performance against state-of-the-art SA and MA MOT schemes via the cluster graph-aware fusion, data-driven Kalman Filtering and wavelet-based temporal refinement modules.

Table \ref{tab:ablation_components} details the contribution of each component to the overall framework. The integration of a data-driven Kalman Filtering tracking pipeline to the MA setting yields poor performance, as the aggregation of multi-vehicle detections increases the FP cause systematic high number of IDS. The DBSCAN approach tends to reduce those tracking errors enhancing sAMOTA and AMOTA by grouping spatially related observations  and forwarding a single representative per cluster to the tracker.
Although single-representative selection may occasionally discard valid detections, incurring a marginal MT reduction, the Graph Laplacian-based fusion module recovers spatial accuracy through differential coordinates and anchor constraints, improving true positive associations. Finally, wavelet-based refinement exploits multi-resolution decomposition of historical trajectories to attenuate short-term positional fluctuations, yielding the most significant gains in sAMOTA and AMOTA and reflecting sustained improvement in long-term trajectory continuity. Although MT and AMOTP do not increase monotonically across all ablation variants, the final configuration achieves the best overall balance across all tracking metrics.

\begin{figure}[ht]
\centering
 \includegraphics[scale=0.35]{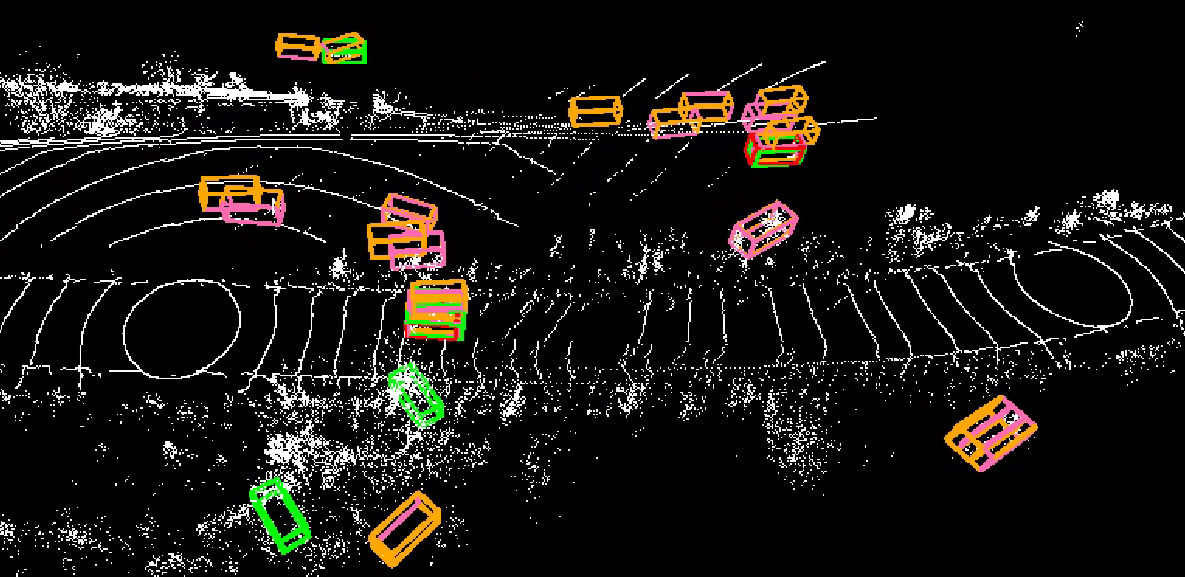}
 \caption{Qualitative results of multi-agent MOT based on assessing tracking accuracy
and precision with GT (red), CDKF-Tracker (green), TSA Graph Lap-CoMOT (pink) and DMSTrack (orange)}
  \label{fig:visuli}
\end{figure}

Figure \ref{fig:visuli} visualizes the ground-truth trajectories (GT), \textbf{CDKF-Track}, TSA-Graph MOT, and DMSTrack trajectories at frame 1 of sequence 0000, denoted in red, green, pink, and orange, respectively. The proposed \textbf{CDKF-Track} provides better spatial alignment with the ground truth and reduces redundant bounding boxes compared with the other multi-agent MOT methods, indicating more consistent and reliable tracking performance on the real-world V2V4Real dataset.

\section{Conclusion}

In this paper, we proposed CDKF-Track, a cluster-aware data-driven Kalman filtering framework for cooperative 3D MOT. The proposed method formulates a cluster graph-aware multi-vehicle fusion observations module, data-driven Kalman filtering, and wavelet-based temporal refinement within a unified tracking-by-detection pipeline. Multi-agent detections are refined through differential coordinates and anchor points, grouped into representative observations, and then associated with trajectories updated by a data-driven Kalman filtering model. Wavelet-based refinement further smooths short-term trajectory fluctuations using historical information. Experimental results on a real-world dataset demonstrate improved cooperative tracking performance over SA and MA MOT baselines, while the ablation study validates the contribution of the proposed components. Future work will investigate multi-source cooperative settings, including vehicle-to-infrastructure scenarios, larger agent networks, and additional real-world and synthetic datasets. 


\bibliographystyle{IEEEtran}
\bibliography{conf}



%



\end{document}